\documentclass{article}

\usepackage{PRIMEarxiv}
\usepackage{amsmath} 
\usepackage[utf8]{inputenc} 
\usepackage[T1]{fontenc}    
\usepackage{hyperref}       
\usepackage{url}            
\usepackage{booktabs}       
\usepackage{amsfonts}       
\usepackage{nicefrac}       
\usepackage{microtype}      
\usepackage{lipsum}
\usepackage{fancyhdr}       
\usepackage{graphicx}       
\graphicspath{{media/}}     

\title{FLASH: Federated Learning-Based LLMs for
Advanced Query Processing in Social Networks
through RAG
}

\author{
  Sai Puppala\\
  School of Computing \\
  Southern Illinois University Carbondale, IL, USA, 62901\\
  \texttt{saimaniteja.puppala@siu.edu} 
  \\
\And
  Ismail Hossain, Md Jahangir Alam, Sajedul Talukder \\
  Computer Science \\
  University of Texas at El Paso, TX, USA, 79968\\
  \texttt{\{ihossain, malam10\}@miners.utep.edu}\\
  \texttt{stalukder@utep.edu}
}
\begin{document}
\maketitle              
\begin{abstract}
Our paper introduces a novel approach to social network information retrieval and user engagement through a personalized chatbot system empowered by Federated Learning GPT. The system is designed to seamlessly aggregate and curate diverse social media data sources, including user posts, multimedia content, and trending news. Leveraging Federated Learning techniques, the GPT model is trained on decentralized data sources to ensure privacy and security while providing personalized insights and recommendations. Users interact with the chatbot through an intuitive interface, accessing tailored information and real-time updates on social media trends and user-generated content. The system's innovative architecture enables efficient processing of input files, parsing and enriching text data with metadata, and generating relevant questions and answers using advanced language models. By facilitating interactive access to a wealth of social network information, this personalized chatbot system represents a significant advancement in social media communication and knowledge dissemination.

\end{abstract}
\section{Introduction}
In a rapidly evolving digital landscape, the fusion of social networks with cutting-edge technology has unlocked unprecedented opportunities to enhance user engagement and information dissemination~\cite{smith2023socialmedia}. One such groundbreaking innovation is the development of personalized social media chatbots empowered by Federated Learning-based GPT (Generative Pre-trained Transformer). This amalgamation offers a paradigm shift in how individuals access social media information, engage with content, and stay updated with the latest trends.

In today's fast-paced world, staying informed about social media trends and events can be overwhelming~\cite{hossain2024visual}. With an abundance of information scattered across user posts, multimedia content, and news articles, navigating this sea of data is a daunting task for both casual users and social media professionals alike. However, imagine if you had a virtual assistant, available 24/7, capable of distilling complex social media data into easily understandable insights, and delivering personalized recommendations tailored to your unique interests and connections.

One major concern is ensuring the privacy and security of user data while utilizing Federated Learning~\cite{puppala2022towards} ~\cite{hossain2023collaborative}. Achieving high performance and accuracy in recommendations generated by social media chatbots is crucial~\cite{abdullah2022chatgpt}. Fine-tuning GPT models or GPT-like decoder-only models~\cite{hossain2024evolve} for social media contexts and ensuring the accuracy of personalized advice require robust validation and optimization processes. Users need to trust the information and recommendations provided by social media chatbots~\cite{biswas2023function}. Ensuring the interpretability of AI-generated insights and transparently communicating the reasoning behind recommendations are essential for fostering trust. Curating relevant and high-quality user-generated content, media, and news articles for users is challenging~\cite{haristiani2020combining}. Encouraging user engagement and adoption of social media chatbots is vital for their success~\cite{ramachandran2019user}. Seamless integration with existing social media platforms and workflows is necessary for the widespread adoption of personalized chatbots~\cite{zumstein2017chatbots}. Addressing these research problems will be essential for realizing the full potential of personalized social media chatbots with Federated Learning-based GPT in enhancing user engagement, information dissemination, and content interaction.

\begin{figure}
\centering
\includegraphics[width=0.7\textwidth]{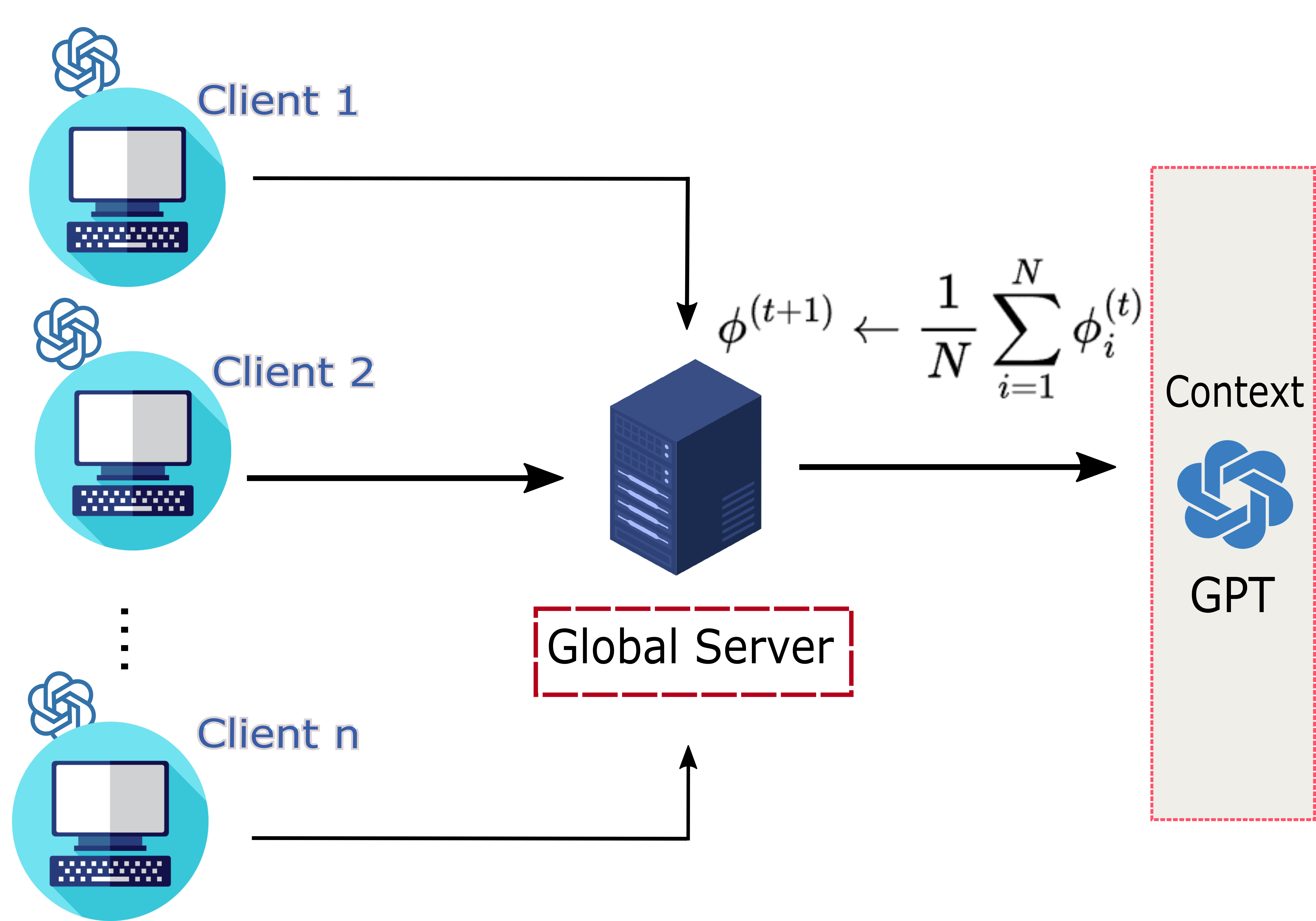}
\caption{Federated learning architecture explaining the workflow of creating social media context-based GPT.}
\label{fig: fed_learn}
\end{figure}

Our research aims to create a secure social media AI system that efficiently curates and delivers diverse social media information through a chatbot interface. Challenges include prioritizing relevant content from various sources, ensuring accuracy, designing an engaging interface, utilizing Federated Learning for privacy, and enhancing user engagement with personalized recommendations (our previous research~\cite{hossain2024socialrec} was developed on personalized post recommendation in social media). Solving these issues is vital for developing a user-friendly chatbot that enhances access to personalized social media information, safeguards privacy, and boosts engagement.

\section{Related Works}
With the rapid advancement of technology, managing the vast array of social network and social media-related information has become increasingly challenging, primarily due to privacy concerns. The sheer volume and diversity of data present a complex landscape that can serve multiple purposes, such as user behavior prediction, addressing social media inquiries, and devising engagement strategies. In their work, Balaji al.~\cite{balaji2021machine} elucidated how machine learning techniques can harness social media data to identify crucial patterns and forecast outcomes. Similarly, Bharadiya et al.~\cite{bharadiya2023machine} employed machine learning methods to anticipate future trends for users based on their past social network interactions.

Given the sensitive nature of social media data, ensuring privacy is paramount when employing data mining and machine learning techniques. Naresh et al.~\cite{naresh2023privacy} explored various privacy-preserving computation methods within the realm of data mining and machine learning to facilitate secure data processing and analysis. Incorporating Federated Learning and cryptography into machine learning frameworks within social media systems, Sravan et al.~\cite{saravanakumar2016privacy} demonstrated how data privacy can be safeguarded, effectively mitigating potential threats like model reconstruction or inversion attacks. Federated learning offers promising solutions for preserving user data privacy across diverse social media platforms, thereby ensuring robust results~\cite{talukder2022novel}~\cite{talukder2022federated}.

As an application of automated systems, machine learning outcomes can manifest as conversational agents equipped with natural language interfaces. Such agents serve to assist users in obtaining tailored responses to their inquiries. wang et al.~\cite{wang2018social} delved into the customization of conversational agents employed within social media, elucidating the methodologies involved in their deployment. In a similar vein, Brand et al.~\cite{brandtzaeg2017people} explored the utilization of chatbot systems in the social media sector, leveraging natural language processing and machine learning methodologies.

\begin{figure}
    \centering
    \includegraphics[width=0.8\textwidth]{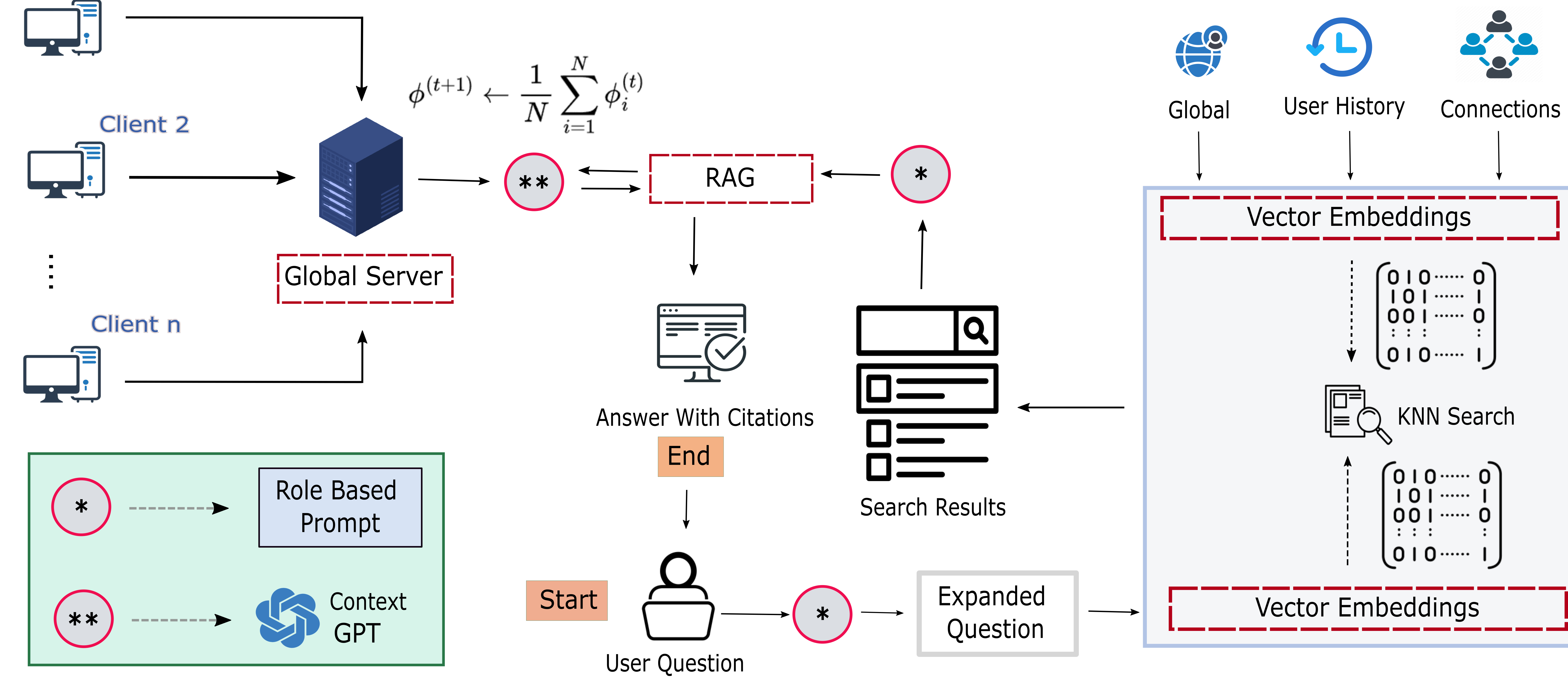}
    \caption{The system architecture illustrates various components. On the leftmost side, the initial module depicts the flow of the federated learning system for generating a context-based GPT. This module outlines the key steps involved in its implementation, from initiation to completion of the workflow. Within the figure, we demonstrate the process of enhancing user queries by supplementing them with pertinent information blocks. These are provided alongside the user's question to the GPT model, prompting it to expand the query based on role-specific cues.}
    \label{fig:system-architecture}
\end{figure}
\section{Methodology}
Our real-time social network chatbot, integrated with Federated Learning approach-based GPT, significantly enhances the efficacy and pace of advancements in social network and practice through its accessible design. Our system aims to furnish user experience and researchers with valuable insights and precise answers through the utilization of large language models like GPT. 

\subsection{Problems with Current State of Art}
Despite the progress made by current state-of-the-art systems, such as ChatGPT and other large language models, certain limitations remain. One key issue is the prolonged processing and analysis time these systems require, which can span months or even years due to the vast amounts of data they handle. Additionally, trustworthiness is a major challenge in applications like ChatGPT. Unlike search engines like Google, which offer users multiple links to help them evaluate the credibility of sources, these systems lack a similar mechanism to ensure trust. This issue is especially critical in areas such as social media, where misinformation can lead to severe consequences. To address this, we have developed a real-time system that prioritizes trustworthiness by providing authentic information and enabling verification through relevant sources. In the following discussion, we will explore the detailed categorization of the various essential components of our system.

\subsection{System Overview}
The introductory section outlines the operational framework of the system, which aggregates data from diverse sources collected using a web crawler from open-source social media. A Python script within a Docker container manages the processing of these data sources. Initially, unstructured text is
parsed, and raw text is extracted into distinct blocks with unique IDs for contextual referencing. These blocks are saved as JSON files enriched with metadata, including embedded mathematical formulas and images for comprehensive context retention. Custom GPT is then utilized to generate relevant
information to the text blocks, employing various prompting techniques. These blocks of information undergo further processing to generate vector embeddings crucial for the system’s functionality, aiding information matching through machine
learning algorithms like Support Vector Machines (SVM). Finally, the processed data are stored in a NoSQL database for superior retrieval performance.

\subsection{Federated Learning}

Federated Learning has recently emerged as a promising architecture that ensures security and privacy for training models. This approach facilitates the training of models across multiple decentralized edge devices or servers that hold local data samples, without requiring the exchange of these samples. In federated learning, each device trains the model locally using its stored data, and only the model updates are sent to a central server or cloud for aggregation. We have employed federated learning to train our custom GPT models, as it ensures the security and privacy of our models.

Here, $\theta^{(t)}$ represents the global model parameters at iteration $t$.
$\eta$ is the learning rate.
$\nabla L(\theta^{(t-1)}, D^{(t)})$ denotes the gradient of the loss function $L$ with respect to the global model parameters using the local dataset $D^{(t)}$.
$\theta_{k}^{(t)}$ represents the local model parameters of client $k$ at iteration $t$.
$\nabla L_k(\theta^{(t)}, D_k)$ denotes the gradient of the loss function $L$ with respect to the local model parameters of client $k$ using its local dataset $D_k$.
The aggregation step computes the updated global model parameters by averaging the local model parameters from all participating clients.

\subsection{Prompt Engineering}
Prompt engineering has become a sophisticated approach for guiding Large Language Models (LLMs) to produce specific results. Its effectiveness and relevance have recently attracted significant interest within the research community. In our study, we have utilized various prompt engineering methods to formulate relevant questions and corresponding answers, specifically tailored to distinct segments of information.

\textbf{Role-Based Prompting}
Role-based prompt engineering is a strategic methodology within machine learning, particularly in natural language processing (NLP) and generative models such as GPT (Generative Pre-trained Transformer) models. In this technique, the model is provided with a customized set of examples or "prompts" that are tailored to specific roles or contexts. These prompts typically consist of a small number of input-output pairs, ranging from a few to several dozen, designed to furnish the model with contextual cues or conditioning information relevant to the specific role it is intended to perform. This context-specific information steers the model's generation of responses or completions.

Once these role-specific prompts are established, the model undergoes fine-tuning based on these examples, facilitating its adaptation and assimilation of the provided prompts. This enables the model to extrapolate to similar tasks or prompts that were not explicitly encountered during training. This approach empowers the model to execute a diverse array of tasks with minimal exemplars, making it particularly advantageous in scenarios where acquiring labeled data is limited or costly.

Let \( X_{\text{role}} = \{x_1, x_2, \ldots, x_m\} \) be the set of role-specific input examples, \( Y_{\text{role}} = \{y_1, y_2, \ldots, y_m\} \) be the set of corresponding output examples, and \newline \( P_{\text{role}} = \{(x_1, y_1), (x_2, y_2), \ldots, (x_m, y_m)\} \) be the set of input-output pairs constituting the role-based prompts. The model is trained to minimize the loss function 
\[
L_{\text{role}}(\theta)
\]
where \( \theta \) represents the model parameters. Fine-tuning involves updating these parameters through gradient descent optimization:
\[
\theta \leftarrow \theta - \eta \cdot \nabla_{\theta} L_{\text{role}}(\theta)
\]
where \(\eta\) is the learning rate.

Through this iterative process, the model learns to generalize from the provided role-specific prompts and adapt to new tasks efficiently. This role-based prompt engineering approach ensures that the model generates responses that are highly relevant and contextually appropriate for the specified role, thereby enhancing the accuracy and utility of the model in practical applications.

\begin{equation}
\begin{aligned}
\text{Global Model Update:} & \quad \theta^{(t)} \leftarrow \theta^{(t-1)} - \eta \nabla L(\theta^{(t-1)}, D^{(t)}) \\
\text{Client Update:} & \quad \theta_{k}^{(t)} \leftarrow \theta^{(t)} - \eta \nabla L_k(\theta^{(t)}, D_k) \\
\text{Aggregation:} & \quad \theta^{(t+1)} \leftarrow \frac{1}{K} \sum\limits_{k=1}^{K} \theta_{k}^{(t)}
\end{aligned}
\end{equation}

\begin{figure}
    \centering
    \includegraphics[width=0.7\textwidth]{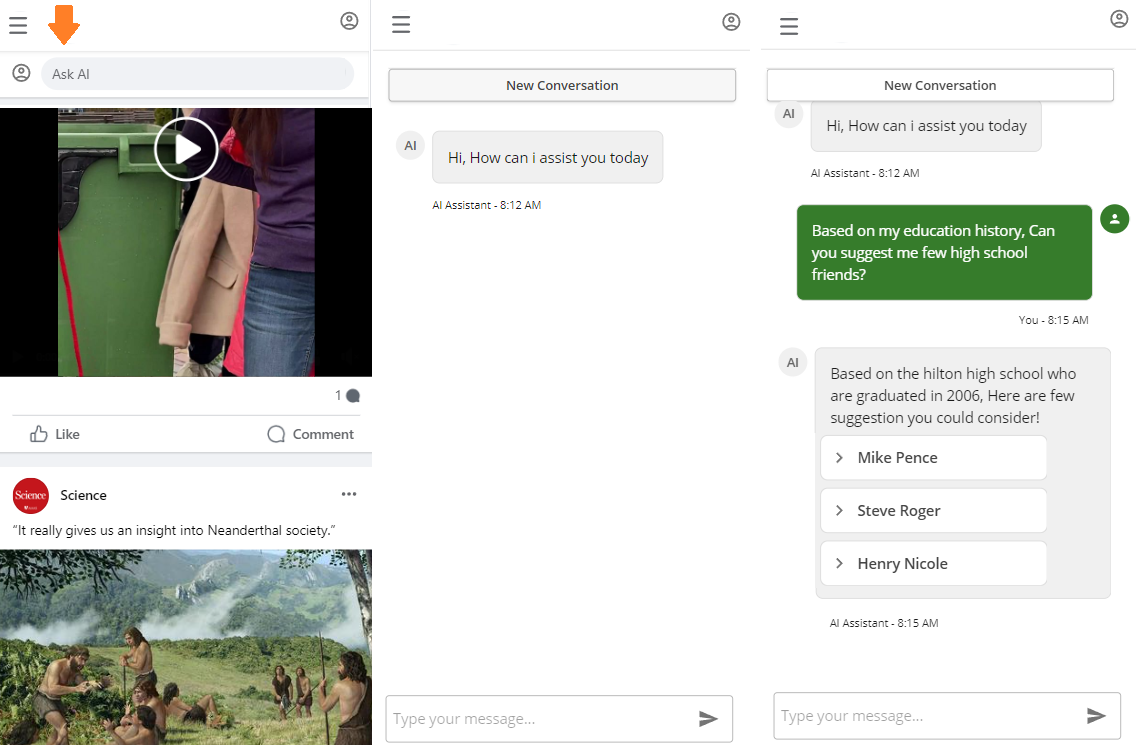}
    \caption{The user interface of the chatbot designed for social media professionals is thoughtfully structured to enhance user experience and streamline interactions. The central feature of the interface is the 'Ask AI' search window, which serves as the primary access point for users to engage with the chatbot. This search window allows users to input their queries or requests, leveraging the chatbot's capabilities to retrieve and generate relevant information or responses.}
    \label{fig:overview_ui}
\end{figure}

\subsection{Vector Embedding}
The vector embeddings operate by leveraging spatial distance measurements to discern closely related questions, facilitating the identification of pertinent information sought by users. This methodology proves instrumental in streamlining the retrieval process within our system. Through the implementation of various algorithms, such as hashing, quantization, or graph-based search, the Vector Embeddings Search Algorithm facilitates an approximate nearest neighbor search. This approach involves locating the closest vector neighbor to a given query, thereby enabling efficient retrieval of relevant information.

Mathematically, the process of identifying the nearest neighbor can be represented as follows: Given a query vector \( q \), and a set of reference vectors \( V = \{v_1, v_2, ..., v_n\} \), the nearest neighbor \( v_{\text{NN}} \) can be found by minimizing the distance metric \( d(q, v_i) \), where \( i \) ranges from 1 to \( n \):

\[
v_{\text{NN}} = \underset{i}{\text{argmin}} \: d(q, v_i)
\]

Vector embeddings constitute a pivotal component of our system, as they encapsulate the essence of data within a sequence of numerical values. Essentially, vector embedding serves as a numerical depiction of the input text furnished to a machine learning algorithm. It encapsulates the semantic relationships and contextual meanings inherent in the words or sentences it processes.

Within our experimental framework, we have harnessed vector embeddings to transform questions generated by large language models into numerical representations. These numerical representations are subsequently utilized to match the input text provided by the user within the front-end application. The comprehensive workflow from inception to execution will be elaborated upon in the ensuing experiment section.

\begin{table}
    \centering
    \caption{The comprehensive statistics, which encompass various clients within the LoRA system implementation, are validated using a range of metrics such as rouge1, rouge2, rouge3, and BLEU. Furthermore, these statistics illustrate the accuracies of the global federated server.}
    \resizebox{\textwidth}{!}{%
\begin{tabular}{|c|c|c|c|c|c|}
\hline
\multicolumn{1}{|l|}{\textbf{Metrics}}            & \textbf{LoRA Federated} & \multicolumn{1}{l|}{\textbf{LoRA Client 1}} & \textbf{LoRA Client 2} & \textbf{LoRA Client 3} & \multicolumn{1}{l|}{\textbf{LoRA Client 4}} \\ \hline

\multicolumn{1}{|l|}{\textit{\textbf{Rouge - 1}}} & 32.383                  & 33.124                                                              & 31.824                 & 32.864                 & 32.372                                                              \\ \hline
 
\textit{\textbf{Rouge - 2}}                                               & 8.245                   & 8.653                                                               & 8.119                  & 7.969                  & 7.932                                                               \\ \hline

\textit{\textbf{Rouge - l}}                                               & 26.804                  & 25.939                                                              & 25.408                 & 25.404                 & 26.819                                                              \\ \hline

\textit{\textbf{BLEU - 4}}                                                & 8.334                   & 8.932                                                               & 8.335                  & 8.467                  & 8.352                                                               \\ \hline
\end{tabular}
} 

        \label{tab:overall-performance}
\end{table}

\begin{table}
    \centering
    \caption{Extensive statistics covering various clients within the P-Tuning-v2 system implementation, validated using diverse metrics including Rouge-1, Rouge-2, Rouge-L, and BLEU. Additionally, the table depicts the accuracies of the global federated server.}
    \label{tab:ml-models-performance}
    \resizebox{\textwidth}{!}{%
    \begin{tabular}{|c|c|c|c|c|c|}
        \hline
        \textbf{Metrics} & \textbf{P-Tuning-v2 Federated} & \textbf{P-Tuning-v2 Client 1} & \textbf{P-Tuning-v2 Client 2} & \textbf{P-Tuning-v2 Client 3} & \textbf{P-Tuning-v2 Client 4} \\ 
        \hline
        \textit{\textbf{Rouge - 1}} & 33.132 & 32.283 & 34.734 & 31.881 & 32.674 \\ 
        \hline
        \textit{\textbf{Rouge - 2}} & 7.899 & 8.879 & 7.126 & 7.268 & 7.121 \\ 
        \hline
         \textit{\textbf{Rouge - l}} & 25.532 & 25.814 & 25.153 & 25.734 & 25.237 \\ 
        \hline
        \textit{\textbf{BLEU - 4}} & 8.245 & 8.392 & 8.736 & 8.829 & 8.981 \\ 
        \hline
    \end{tabular}
     } 
\end{table}

\section{Experiments}
Our proposed architecture is composed of three distinct modules. The first module is dedicated to the development of a customized GPT model, specifically tailored to social networks and social media. Data for this module were collected using web crawlers, with the primary datasets sourced from open-source social platforms. The second module integrates user history, user connections, and global population-related data. The third module encompasses the AI user interface, which is comprehensively illustrated in Figure~\ref{fig:overview_ui}. The overarching goal of our system is to address the challenges associated with context-based generative models and to provide more relevant information to end users. These modules have been systematically organized according to their respective functionalities and have been successfully implemented in a concurrent manner.

\subsection{Context GPT with Federated Learning}
Training our GPT model with social media data is imperative, given that existing generative models available online are predominantly generic and lack contextual depth. To address this, we initiated the training of our customized GPT model using a distributed computing architecture comprising four client or worker nodes. Furthermore, we implemented federated learning to bolster security and privacy, a process that is detailed comprehensively in the methodology section.

\textbf{\textit{Workflow:}} The process initiates by supplying unstructured text as input data to four client nodes, which commence training using the GPT-2 base model. Upon completion of training and validation, these client nodes transmit their custom-trained GPT models to a central global server. The global server employs the federated averaging (fed-avg) algorithm to aggregate these models, forming a collaborative GPT model. This collaborative model is subsequently utilized in phase two to answer user questions based on provided prompts. 

To enhance training efficiency and address the challenges of processing extensive datasets, we integrate popular Parameter Efficient Fine-Tuning (PEFT) strategies, such as LoRA. These strategies markedly reduce the number of trainable parameters and processing times. However, the transportation of large GPT models within a federated learning framework poses significant network challenges. To mitigate these issues, we have incorporated a checkpoint strategy into the system, which helps reduce costs and facilitates smoother model aggregation.

These strategies and implementations represent a comprehensive approach to federated learning, effectively addressing both training efficiency and network constraints. By combining techniques like PEFT with federated averaging and checkpoint strategies, our system optimizes model training and aggregation processes. This approach advances the development of near real-time model production while managing the complexities inherent in distributed computing and federated learning environments.

\textbf{\textit{Network Optimization:} } A checkpoint strategy involves saving model checkpoints at regular intervals during training. By comparing these checkpoints from different models, we can identify which parameters have changed between them. Only these differentials are transmitted to the global server, where the parameter changes are evaluated. Managing the storage and aggregation of these large models on the global server presents a significant challenge. In our scenario, with each of the four client machines producing a custom GPT model of 0.5 gigabytes, the total model size on the global server amounts to 2GB.

To address this, we have adopted quantization, a technique that significantly reduces the overall model size. After collecting all the GPT models generated by the client machines, the global aggregation process leverages quantization to prepare the model for use. This approach not only alleviates the storage and transmission burden but also enhances the efficiency of the model aggregation process.

\textbf{\textit{Context GPT Evaluation:}} For the evaluation of the models generated by the client machines, we employed metric scales such as Rouge-1, Rouge-2, Rouge-L, and BLEU-4. These metrics assess the quality of the generated output by evaluating the uni-grams, bi-grams, and N-grams of word sequences in comparison to the original sentences. Table~\ref{tab:overall-performance} provides a detailed breakdown of these evaluation metrics for both the global server and the four client machines, highlighting the overall performance across different measurement scales. Additionally, Table~\ref{tab:ml-models-performance} showcases the performance levels achieved by implementing the P-Tuning PEFT strategy, incorporating the latest version for enhanced accuracy. Furthermore, Table~\ref{tab:params-performance} offers insights into the training parameters and sizes of the GPT-generated models across the client machines, illustrating the various trainable model sizes and parameters.

\begin{table}
  \centering
  \caption{The detailed statistics encompassing various trainable model parameters and sizes are essential components of our federated learning implementation aimed at constructing a context-based GPT.}
\begin{tabular}{|l|c|c|}
\hline

\textbf{Methods}                                                            & \textbf{Model Size (MB)} & \multicolumn{1}{l|}{\textbf{Param Percent (\%)}} \\ \hline
 
\textit{\textbf{Global Sever}}                                              & 6173                     & -                                                                        \\ \hline

\multicolumn{1}{|c|}{\textit{\textbf{LoRA}}}        & 3.6                      & 0.058                                                                    \\ \hline

\multicolumn{1}{|c|}{\textit{\textbf{P-Tuning-V2}}} & 29.3                     & 0.475                                                                    \\ \hline
 
\multicolumn{1}{|c|}{\textit{\textbf{Checkpoint}}}  & 6.2                      & 0.116                                                                    \\ \hline

\textit{\textbf{Full Tuning}}                                               & 6173                     & 100                                                                      \\ \hline
\end{tabular}
\label{tab:params-performance}
\end{table}

\subsection{Information Retrieval System}
This module is crucial for understanding and establishing a vector space for user history, user connections, and global social media data. It is important to highlight that the data used by client nodes in federated learning for context GPT generation is entirely distinct from the data processed in the second module. 

The primary objective of this module is to accurately identify and retrieve the appropriate blocks of information based on the user's query. For instance, when a user poses a question, we utilize a role-based prompting mechanism, as detailed in the Methodology section, to expand the query and enrich it with additional contextual information. This expansion process enables the system to better understand the user's intent and effectively locate the most relevant information blocks. 

\textbf{\textit{Expanded Question:}} To enhance the contextual understanding of user queries, our approach involves the systematic expansion of these questions. This is achieved by incorporating headers from information blocks, which have been previously collected using web crawlers and stored in a structured database. These headers serve as concise summaries of the information blocks' content. When a user submits a query, we employ a role-based prompting mechanism that utilizes these headers to generate a context-rich prompt for a context-aware GPT model. The model, trained to assimilate and process contextual information, responds by producing an elaborated version of the original user question, enriched with additional context. This enhanced query formulation process significantly aids in accurately identifying and retrieving the most relevant information blocks during an open search, thereby improving the precision and relevance of the search results.

\begin{figure}
    \centering
    \includegraphics[width=0.8\textwidth]{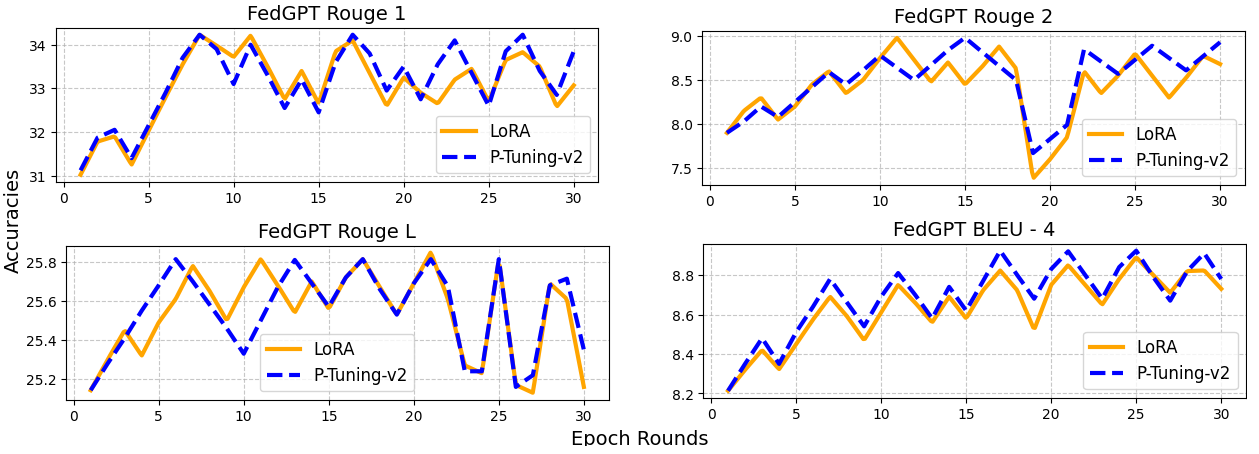}
    \caption{The four figures above depict varying accuracy levels for FedGPT across different metric scales. Blue represents p-Tuning, while orange represents LoRA. The y-axis indicates accuracies, and the x-axis represents the number of training rounds conducted.}
    \label{fig:stats}
\end{figure}

\textbf{\textit{Vector Embeddings:}} Following the expansion of the user's question, vector embeddings are generated for the enriched query. These embeddings are then utilized to conduct a search across all the vector embeddings stored in the open search system, which represent the information blocks collected by web crawlers. To identify the most relevant blocks of information, we employ a Support Vector Machine (SVM) algorithm. The SVM algorithm efficiently processes the vector embeddings to determine the nearest and most pertinent information blocks, thereby enhancing the accuracy and relevance of the search results.

The vector embeddings serve as a foundational element in the search algorithm employed when end-users or customers submit queries through the front-end interface. These embeddings, which encapsulate the semantic context of the input data, enable the system to match user queries with the most relevant content swiftly and accurately.

Furthermore, in addition to storing these embeddings, the system also generates keywords from the user queries. This keyword generation is integral for implementing smart filtering mechanisms. By extracting and leveraging these keywords, the system enhances the relevance and precision of the information retrieved, ensuring that users receive the most pertinent and contextually appropriate responses.

The integration of vector embeddings and OpenSearch, coupled with keyword generation for smart filtering, represents a robust and efficient approach to information retrieval in social media applications. This methodology enhances the system's ability to deliver highly relevant and accurate responses to user queries, thereby improving user satisfaction and engagement. The strategic use of advanced technologies like OpenSearch and vector databases underscores the importance of leveraging state-of-the-art tools to meet the demands of real-time, high-performance search algorithms in dynamic and data-rich environments.

\textbf{\textit{Prompt Engineering: }} The subsequent step entails utilizing context-specific GPT with a role-based prompt engineering technique. Role-based prompt engineering is a strategic approach that involves crafting prompts tailored to the specific roles or contexts in which the language model is deployed. This technique ensures that the model generates responses that are highly relevant, accurate, and contextually appropriate for the intended application.

In the third step, the information blocks identified through the SVM search using vector embeddings are provided as input to the context-aware GPT model, along with the original user query. The context GPT model then evaluates these information blocks and synthesizes a response to the user's question, ensuring that the answer is informed by the most relevant and contextually appropriate data.

\section{Conclusion}
In conclusion, our paper introduces an innovative Federated Learning (FL) approach integrated with GPT, specifically tailored for social networks and social media applications. Our system enables secure model training and communication over decentralized networks. By structuring our system into multiple modules, we facilitate concurrent task execution, resulting in more real-time outcomes compared to traditional systems. Unlike existing architectures that obscure the data sources, our approach provides transparency regarding data origins and allows users to view or set specific contexts. We assert that our proposed system delivers superior insights compared to current state-of-the-art methods, particularly in terms of the accuracy and relevance of the information provided.

\end{document}